\documentclass[letterpaper,man,floatsintext]{apa7}

\usepackage[utf8]{inputenc}
\usepackage{url}

\usepackage[numbers,square,sort&compress]{natbib}

\usepackage{enumitem}
\usepackage{setspace}
\usepackage{hyperref}
\hypersetup{
  colorlinks=true,
  linkcolor=blue,
  citecolor=black,
  urlcolor=black
}

\makeatletter
\newif\ifusecolor
\usecolorfalse  

\ifusecolor
\else
  \renewcommand{\color}[1]{\relax}
  
\fi
\makeatother

\title{The Homogenizing Effect of Large Language Models on Human Expression and Thought}
\author{
  Zhivar Sourati\textsuperscript{1,2},
  Alireza S. Ziabari\textsuperscript{1,2},
  Morteza Dehghani\textsuperscript{1,2,3}
}

\affiliation{
  \textsuperscript{1}Department of Computer Science\\
  \textsuperscript{2}Center for Computational Language Sciences\
  \textsuperscript{3}Department of Psychology, University of Southern California
}

\shorttitle{Homogenizing Effect of LLMs on Cognitive Diversity}

\abstract{Cognitive diversity, reflected in variations of language, perspective, and reasoning, is essential to creativity and collective intelligence. This diversity is rich and grounded in culture, history, and individual experience. Yet as large language models (LLMs) become deeply embedded in people's lives, they risk standardizing language and reasoning. {\color{blue}We synthesize evidence across linguistics, psychology, cognitive science, and computer science to show how LLMs reflect and reinforce dominant styles while marginalizing alternative voices and reasoning strategies.} We examine how their design and widespread use contribute to this effect by mirroring patterns in their training data and amplifying convergence as all people increasingly rely on the same models across contexts. Unchecked, this homogenization risks flattening the cognitive landscapes that drive collective intelligence and adaptability.}

\begin{document}

\maketitle

\section{When LLMs Meet Human Diversity in Expression and Thought}

Cognitive diversity, intertwined and manifested through varied linguistic expressions, is essential to the adaptability, creativity, and overall effective functioning of complex societies \citep{page2008difference}. {\color{blue} 
Such variation, while might be expressed through stylistic differences, reflects deeper cognitive \citep{evans2009myth,majid2004can} and sociocultural differences \citep{eckert2012three}, and vitally, plays a critical role in sustaining the \textbf{epistemic} (see Glossary) and problem-solving capacities of human groups \citep{hong2004groups}. 
This value of pluralism is rooted in the long-held principle that sound judgment requires exposure to varied thought. As John Stuart Mill argued, ``the only way in which a human being can make some approach to knowing the whole of a subject, is by hearing what can be said about it by persons of every variety of opinion, and studying all modes in which it can be looked at by every character of mind. No wise man ever acquired his wisdom in any mode but this'' \citep{mill2001liberty}. When preserved, such distinctions support innovation, prevent \textbf{epistemic collapse} (see Glossary), and enhance the operational efficacy of collective systems \citep{solomon2006groupthink,page2008difference}.}

This diversity has emerged organically from the coexistence of individuals with distinct backgrounds, linguistic repertoires, and value systems \citep{labov1972social,evans2009myth}. Yet, the increasing reach of global communication technologies, while enabling unprecedented knowledge sharing and connection, has also contributed to a gradual contraction of such linguistic and cognitive variation \citep{harmon2010index,alsaleh2024impact,cooperrider2019happens}. Among these technologies, Large Language Models (LLMs) have emerged as especially influential, becoming deeply integrated not only within digital infrastructure but also as fundamental components shaping how we interact with technology and with each other \citep{handa2025economic}. 

Extending beyond traditional language-based applications such as summarization tools \citep{zhang2024benchmarking}, LLMs are now involved in tasks once reserved for human, such as sociocognitive modeling \citep{kosinski2023theory}, psychological simulation \citep{park2023generative}, and even experimentally in place of human participants \citep{aher2023using,dillion2023can}, which expands the scope of what it means for LLMs to represent human diversity. This integration makes it critical to examine whether these models preserve human diversity or instead enforce a form of cognitive and linguistic homogenization. 
{\color{blue}
Although excessive diversity can introduce costs related to coordination, communication, and coherence, and some degree of standardization may help mitigate these challenges \citep{wang2022inverted}, the risks inherent in this standardization are substantial:
}
homogenized generations may constrain public discourse, reduce the visibility of marginal linguistic forms, and reinforce dominant reasoning templates. They may also suppress the kinds of idiosyncratic language use that signal individual traits or group-specific perspectives \citep{sourati2025shrinking}. 
In complex reasoning tasks, the widespread adoption of \textbf{chain-of-thought prompting} (\citealp{wei2022chain}; see Glossary), optimized for linear and explicit inference, may disincentivize more abstract or intuitive reasoning styles that are harder to model yet crucial for flexible problem-solving \citep{sui2025stop,ziabari2025reasoning}. These shifts raise broader concerns: that LLMs, if uncritically integrated, may shape not only how we write but how we think.

{\color{blue}
Anxieties about technology's influence on language and cognition have a long history, starting with Plato's Phaedrus and his concern that writing would weaken memory. Modern research echoes this, showing the Internet enables people to offload knowledge externally, which can inflate confidence in what they know \citep{fisher2015searching}. While such externalization has been framed as potentially freeing mental resources for creativity and problem-solving \citep{clark1998extended}, this benefit is less likely to manifest with the widespread adoption of LLMs.

Unlike prior technologies that primarily aided storage or retrieval, LLMs act as fluent co-reasoners, and at times, standalone ones, participating in writing, problem-solving, and perspective-taking, thereby externalizing not only memory but the articulation and justification of thought. As Clark \& Chalmers \citep{clark1998extended} argue, ``Language, thus construed, is not a mirror of our inner states but a complement to them. It serves as a tool whose role is to extend cognition in ways that on-board devices cannot.'' When that linguistic interface is mediated by systems capable of generating reasoning and perspective, much of human cognition risks being relocated outside the individual mind.

Hence, the effect of LLMs, though similar in nature to earlier cognitive extensions, differs profoundly in both function and scale. Earlier systems, such as schools, books, and search engines, while capable of promoting cultural uniformities, mainly disseminate knowledge or teach reasoning frameworks that individuals must internalize and apply. This internalization is an active, personalized process through which frameworks are adapted, integrated with pre-existing knowledge, and eventually transformed into unique cognitive strategies that are deployed as needed, often at later times. LLMs, by contrast, generate complete reasoning and articulation processes themselves on users' behalf. The same few models now generate text for hundreds of millions of users, with adoption rates surpassing those of any previous digital technology \citep{bick2024rapid}. Their use has expanded especially rapidly in non-work contexts \citep{chatterji2025people} and remains concentrated around a small number of dominant systems, such as ChatGPT \citep{wiggers2025chatbots}. This convergence, coupled with their linguistic fluency that can lead users to overtrust their responses \citep{steyvers2025large}, collapses the boundary between external tool and internal cognition, standardizing not only the information people access but also the very ways they articulate, justify, and reason through ideas.
}

\begin{figure}[ht]
    \centering
    \includegraphics[width=1\linewidth]{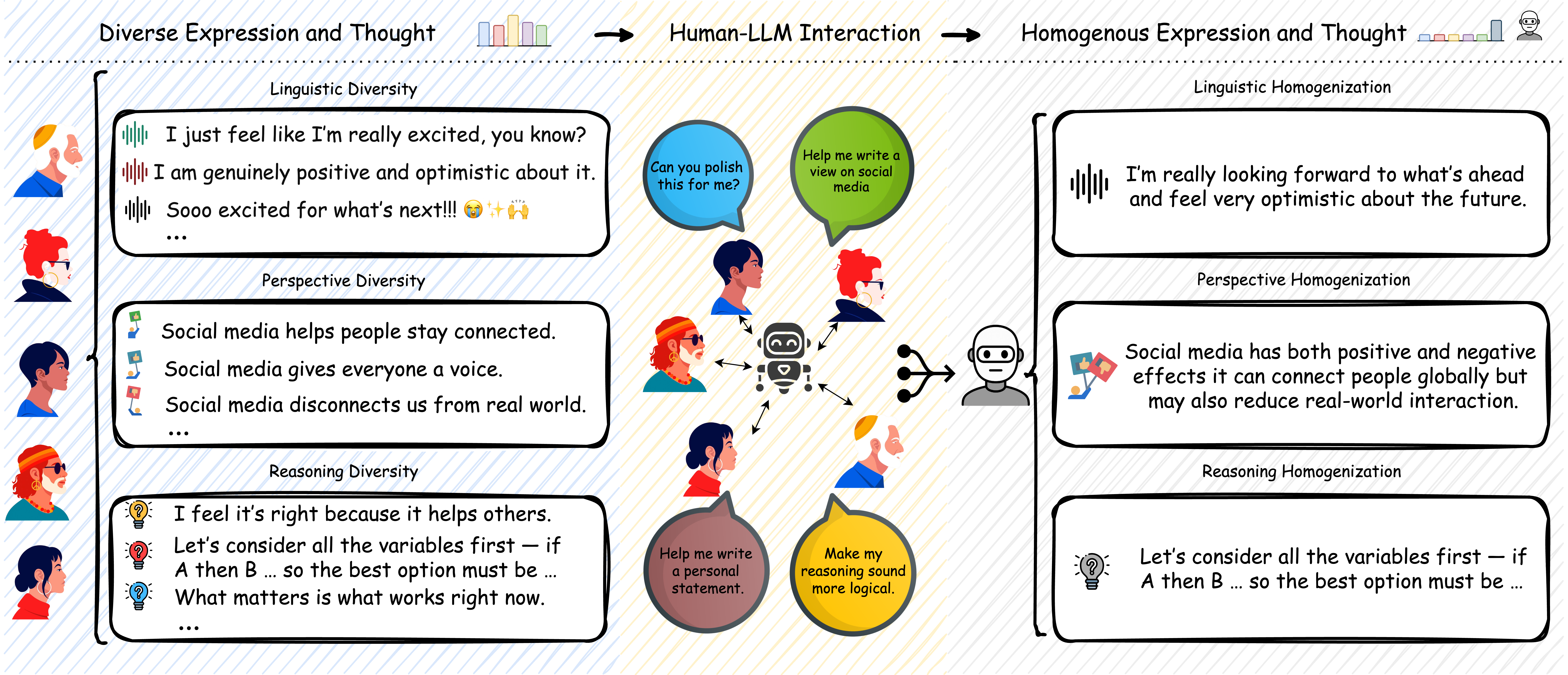}
    \caption{Individuals differ in how they write, reason, and view the world. When these differences are mediated by the same LLM, their distinct linguistic, perspectival, and reasoning signals become homogenized, producing standardized expressions and thoughts across users.}
    \label{fig:main-figure}
\end{figure}

LLMs thus present a paradox for cognitive science: as they become a powerful tool for modeling human thought \citep{strachan2024testing}, they also risk flattening the very cognitive diversity that cognitive science studies \citep{boogert2018measuring}. With growing concern across linguistics, computer science, psychology, and cognitive science more broadly about the homogenizing effects of LLMs, we bring these perspectives into conversation. We discuss diversity across three key dimensions: stylistic variation in language, perspective, and reasoning strategies (see \autoref{fig:main-figure}). {\color{blue}
By bridging disciplinary boundaries, we aim to provide a more integrated understanding of how LLMs interact with human diversity, both reflecting and influencing variation in human reasoning and expression, and argue for greater incorporation of human-grounded diversity in these dimensions within the models.
}

\section{Language Models, Prediction, and the Loss of Diversity}

Language modeling is central to modern artificial intelligence as it provides the foundation for systems like GPT-4 \citep{achiam2023gpt} and Gemini \citep{team2023gemini} that understand, generate, and interact through natural language \citep{zhao2023survey}. 
At their core, these models operate by predicting the next token given a preceding context, a goal shared with earlier approaches like $n$-gram models that estimated token probabilities using fixed-size word windows and explicit statistical counts from large corpora \citep{mitchell1999machine}. LLMs extend this approach at a much larger scale, trained on massive datasets with billions of parameters, supported by unprecedented levels of compute and architectural optimization \citep{grattafiori2024llama}. Despite these advances, their fundamental objective remains the same: mastering the statistical regularities of language, which remains the dominant driver of model behavior \citep{lin2023unlocking} even after the application of LLM alignment methods such as \textbf{supervised fine-tuning} (\citealp{wei2021finetuned,ghosh2024closer}; see Glossary) and preference-based alignment tuning via \textbf{reinforcement learning from human feedback} (RLHF; \citealp{bai2022training,wolf2023fundamental}; see Glossary).

Yet these advances come with inherent limitations. Because LLMs are trained to capture and reproduce the statistical regularities of their input data, which often overrepresent dominant languages and ideologies \citep{schut2025multilingual,buyl2024large}, their outputs tend to mirror a narrow and skewed slice of human experience. This limitation arises not only from biased training corpora but also gets amplified through the training process itself \citep{wang2024bias}, favoring patterns that are frequent and easily generalizable while smoothing over minority representations \citep{hall2022systematic}.
{\color{blue}
At scale, what begins as statistical pattern learning, though capable of generalization beyond mere mimicry \citep{mccoy2023much,misra2024language}, becomes a generative force that privileges central tendencies while marginalizing rare expressions, alternative reasoning styles, and culturally specific voices.
} 
The consequence is not just a convergence in surface-level linguistic form, but a narrowing of the conceptual space in which models write, speak, and reason \citep{schut2025multilingual}.

Critically, this narrowing does not trend toward a neutral center but toward a historically uneven one, shaped by the norms, values, and perspectives of English-speaking, Global North, and socioeconomically advantaged populations \citep{alvero2024large,hartmann2023political}. This has tangible consequences: 
{\color{blue}
when prompted for opinions or expressive writing, LLMs tend to reproduce mainstream, institutionally validated perspectives and writing styles that mirror those of western, liberal, high-income, highly educated males, creating an illusion of consensus that frames these norms as the default standard of clarity or intelligence while muting alternative worldviews and culturally grounded forms of expression \citep{alvero2024large,sourati2025shrinking}. This misrepresentation persists even when models are explicitly prompted to assume a specific identity, often resulting in LLM personas that reflect out-group stereotypes rather than authentic in-group representations \citep{wang2025large}. For instance, when asked for a person with impaired vision's thoughts on immigration, one model responded: ``While I may not be able to visually observe the nuances of the US-Mexican border or read statistics, I believe...'' \citep{wang2025large}. This misportrayal is a symptom of a larger problem where LLMs flatten demographic groups by neglecting heterogeneity and, through identity prompting, reduce identities to fixed, \textbf{essentialized representations of identity} (see Glossary). Consequently, what is produced is not a prototype of the group's varied experience, but a mere caricature.
}

This representational imbalance is not static; it becomes an active, homogenizing force through a recursive feedback loop \citep{wyllie2024fairness}. As individuals increasingly turn to LLMs for writing, problem-solving, and conceptual exploration, the models' outputs, already favoring common linguistic and conceptual patterns, is reabsorbed into human discourse, and begins to shape the users' own expression and reasoning \citep{jakesch2023co} and, in turn, influencing the data used to train future models, transforming homogenization from a passive bias into a structurally reinforced influence. 


In the following subsections, we examine the homogenizing effects of LLMs and how this recursive influence manifests across three interrelated domains: language, perspective, and reasoning.

\section{Language Diversity}
\label{sec:language-diversity}

Across time and geography, human groups have developed distinct linguistic systems and cultural norms \citep{nichols1992linguistic,atari2023morality}, shaped by environmental conditions and social structures \citep{nolle2020language,eckert2012three}. These dynamics are encoded and preserved through language, rich symbolic systems that carry not only the content of communication but also people's underlying values and identities \citep{boyd2015values,bamman2014gender}. 
{\color{blue}
For example, people categorized as extroverts tend to use more words related to humans, social processes, and family in personal essays about past experiences, reflecting a greater focus on interpersonal connection and social engagement \citep{hirsh2009personality}. 
}
Language, in this sense, serves as a powerful medium for expressing cross-language and within-language individual and group differences \citep{pennebaker2011secret,boyd2015values}, not just through semantic content but also through subtle stylistic \citep{pennebaker2011secret} and structural cues \citep{boghrati2018conversation}.

{\color{blue}
Efforts to enhance linguistic diversity have long been central to \textbf{Natural Language Processing} (NLP; see Glossary), preceding the emergence of large language models. Work in this area primarily focused on improving diversity in language generation, aiming to make machine-produced text more informative, engaging, and natural across tasks such as summarization \citep{li2009enhancing}, translation \citep{gimpel2013systematic}, and more broadly, dialogue generation \citep{li2015diversity}. 
}

{\color{blue}
Concurrently, fields such as computational sociolinguistics and authorship profiling have focused on exploring how language reflects speakers' underlying backgrounds, and introduced various methods to identify linguistic signatures of social position, and individual traits \citep{pennebaker2011secret,nguyen2016computational}. For instance, researchers have analyzed congressional speeches to predict speakers' age and gender from linguistic cues in their public addresses \citep{ziabari2024reinforced}. Yet, the primary objective of these works was analytical rather than generative, as until recently, most NLP research focused on producing engaging conversations with surface-level linguistic diversity.
}

However, as LLMs have mastered surface-level fluency and natural linguistic variation \citep{guo2024benchmarking}, a central question emerges: do they exhibit human-like patterns of linguistic variation, i.e., do they reflect and preserve socially meaningful forms of variation and preserve links to speaker traits? Existing evidence suggests they do not. 
{\color{blue}
Recent research shows that when LLMs are used to polish various forms of writing, ranging from Reddit posts and news articles to academic abstracts and personal essays, the resulting texts converge in writing complexity, diminishing the predictability of author characteristics such as political affiliation, personality, gender, or age, and even weakening well-established associations like the link between ``big-word'' usage (words with complex structures of seven letters or longer, that may indicate intellectual complexity or formality) and the author's openness to experience \citep{sourati2025shrinking}. Likewise, studies generating college admission essays entirely with LLMs find that the resulting texts exhibit high semantic and lexical similarity across samples, reflecting a narrowing of expressive space \citep{lee2025poor}. Importantly, this homogenization persists despite interventions such as \textbf{temperature scaling} (see Glossary) or different prompting techniques that simulate different author identities or personas, which are commonly used to induce stylistic diversity \citep{martinez2024beware,guo2024benchmarking,lee2025poor}.
}


This bias appears to worsen as models are increasingly trained on synthetic, model-generated data \citep{guo2023curious}, and is further amplified by the adoption of reinforcement learning (RL), particularly RLHF, to enhance perceived qualities such as helpfulness \citep{bai2022training} and reasoning capabilities \citep{guo2025deepseek}. While these RL-based techniques have substantially improved reasoning compared to fine-tuned models \citep{chu2025sft}, they have also been shown to reduce stylistic and expressive variability \citep{kirk2023understanding}. This effect is reinforced by the quality–novelty trade-off, where methods that promote novelty, often compromise coherence \citep{zhang2020trading}, suggesting that continued optimization for quality and performance may further diminish linguistic and stylistic diversity.

To address these issues, researchers have proposed several strategies. Prompting-based methods, applied at inference time, aim to enhance output diversity by modifying how prompts are phrased or conditioned \citep{chu2024exploring,kambhatla2025measuring}. 
{\color{blue}
For instance, studies using persona-based prompts show that coarse-grained persona conditioning, combined with adjustments to output length, can maximize lexical variation in model responses \citep{kambhatla2025measuring}. Training-based approaches instead modify the LLMs' learning process to optimize for semantic and stylistic diversity \citep{mai2024improving,chung2025modifying}. For example, researchers have fine-tuned models on datasets with multiple valid responses per prompt \citep{mai2024improving} or introduced diversity-aware weighting in preference optimization to reward rare, high-quality generations \citep{chung2025modifying}. Although promising, evidence of homogenization in sociolinguistic contexts calls for closer evaluation of whether these strategies foster genuine, context-grounded diversity or merely superficial variation.
}

Vitally, the concerns regarding the linguistic diversity of LLMs extend beyond representation alone, as these models actively shape how language is used and evolves. As they become embedded in everyday writing (e.g., composing emails), while supporting more efficient and polished communication \citep{hohenstein2023artificial}, they also tends to promote uniform styles that mask authentic voices and reduce variation in tone, culture, and identity \citep{alvero2024large,holliday2025gender} even for those merely engaging with AI-generated text \citep{toma2014towards}.

{\color{blue}
At first glance, the homogenization of language, along with the loss of lexical cues linking writing to individual identity, may appear beneficial for protecting privacy as it reduces risks of misuse such as surveillance or discrimination \citep{tufekci2014engineering}. But it is equally important to recognize that this process also erases valuable linguistic markers long used across sociolinguistics, psychology, and mental health research. For example, people at risk of Alzheimer's disease exhibit early linguistic signs such as telegraphic speech (simplified phrases lacking grammar and function words, often missing determiners like ``the'' or ``a,'' auxiliaries like ``is'' or ``are,'' and even entire subjects), as well as repetitiveness and misspellings, patterns reflecting a decline in grammatical structure and language complexity \citep{eyigoz2020linguistic}. If such distinctive markers are standardized by LLM-assisted polishing, critical early indicators for diagnosis and intervention may be lost. Moreover, the writing styles that LLMs incorporate are not neutral but a source of bias themselves, reproducing dominant expressive norms while eroding minority and underrepresented voices \citep{alvero2024large,huang2024translating}. 
}

\section{Perspectival Diversity}

Individuals vary not only in how they express themselves but also in their perspectives, beliefs, and values \citep{aglozo2025cross}.
In language, these differences are reflected in patterns of how people take stances \citep{kuccuk2020stance} or frame issues \citep{sagi2014measuring}, widely studied in computational analyses of text \citep{kennedy2022text}. 
{\color{blue}
For instance, analyses of U.S. Senate speeches reveal that Democrats and Republicans frame the issue of abortion through distinct moral dimensions, Democrats emphasizing fairness and rights, while Republicans focus more on purity and sexual morality, mirroring broader ideological divides in moral reasoning \citep{sagi2014measuring}.
}

{\color{blue}
With the rise of LLMs, this domain of variation has become especially salient, as these models are increasingly deployed in contexts involving perspective gathering and open-ended writing tasks \citep{argyle2023out}. Recent work shows they can simulate public opinion, for example, on global warming, by generating synthetic survey responses conditioned on demographics and personal concern, capturing belief patterns that mirror human data \citep{lee2024can}. Hence, a pressing question arises: do they preserve the pluralism of human perspectives, or do they default to a narrow, socially normative median?
}

Multiple studies have shown that LLMs tend to reflect characteristic of western, educated, industrialized, rich, and democratic societies (WEIRD) in their perspectives \citep{abdurahman2024perils,atari2023humans,durmus2023towards}. 
{\color{blue}
Using the World Values Survey \citep{haerpfer2020wvs} and the Moral Foundations Questionnaire-2 (MFQ-2; \citealp{atari2023morality}) researchers tested GPT-3.5 across 1000 runs with a temperature of 1.0 to maximize output variance and compared the results with human responses from diverse cultural backgrounds, finding that LLM outputs not only exhibited substantially less variance than human data but were also more closely aligned with response patterns characteristic of WEIRD societies, showing limited variability and weak representation of non-WEIRD perspectives \citep{abdurahman2024perils,atari2023humans}. While models can simulate diverse viewpoints when explicitly prompted, such as by adjusting generation parameters, incorporating identity-coded instructions \citep{li2024can}, or translating prompts to a target language \citep{wang2025multilingual}, these interventions have often fallen short of aligning outputs with the actual distribution of perspectives within the referenced groups \citep{durmus2023towards,santurkar2023whose}, capture merely the socially ``correct'' \citep{park2024diminished} or the mean of the distribution \citep{abdurahman2024perils}, and in some cases, even reproduce out-group stereotypes or misrepresentations of the specified populations \citep{wang2025large}. 
}

{\color{blue}
Similar to efforts aimed at enhancing lexical diversity in LLM generations, recent research has also sought to increase perspective diversity through both prompting and fine-tuning approaches \citep{hayati2023far,hu2024debate}. For instance, researchers have introduced a debate-based framework in which multiple models interact to produce arguments that are broader in scope and more balanced across viewpoints \citep{hu2024debate}. Further through fine-tuning, reinforcement learning frameworks have been adapted to mitigate ``preference collapse'' by adjusting reward functions to account for underrepresented or minority preferences, balancing performance with diversity \citep{xiao2024algorithmic}. Yet it remains unclear whether these strategies truly embody human-aligned pluralism of perspectives and contextual depth, highlighting the need for more systematic evaluation.
}

{\color{blue}
This power to simulate and frame perspectives does not remain contained within the model; as LLMs become integrated into daily lives, they begin to influence how individuals perceive and frame the world, narrowing perspectives that are naturally rooted in lived experiences \citep{hwang202580}, environmental, and social contexts \citep{lubart2001models,nijstad2002cognitive}. 
Studies show that when people remember events together, they tend to align their memories with others in the group, reinforcing shared details while forgetting those left unmentioned \citep{coman2015social}. LLMs, however, can amplify this dynamic on a global scale: by exposing millions of users to the same suggestions and perspectives, they foster similar patterns of recall and association, promoting convergence in what people collectively remember and express.
}

One setting where this influence is particularly consequential is co-writing with LLM-based assistants, as people increasingly rely on them even for open-ended survey responses about their beliefs and behaviors \citep{zhang2025generative}, and this, in turn, shapes how users frame and articulate their own perspectives \citep{jakesch2023co,agarwal2025ai}.
{\color{blue}
This effect is evident in studies showing that participants who co-wrote with opinionated language models, engineered to frame social media positively or negatively, tended to mirror the model's stance in their writing and even shifted their own opinions in subsequent attitude surveys \citep{jakesch2023co}.
}
These findings raise broader concerns about persuasive influence, as even subtle interaction with LLMs can lead users to adopt the model's framing without awareness \citep{rashotte2007social}.

\section{Reasoning Diversity}

Beyond language and perspective, another consequential form of diversity at stake is reasoning diversity, the varied ways people reason, solve problems, and generate new ideas, which arises from cultural and individual differences. For example, Native American (Menominee) children often group animals based on ecological relationships, such as shared habitat or interdependence, whereas European-American children tend to rely on taxonomic categories, reflecting broader differences in how people make sense of the world \citep{medin2004native,medin2007culture}. Across disciplines, variation in reasoning is recognized as a key driver of collective strength \citep{nijstad2002cognitive,page2008difference}. Groups composed of individuals who reason differently, using distinct heuristics and problem representations, consistently outperform groups made up of the highest-ability individuals \citep{hong2004groups}, and innovation and intellectual breakthroughs often emerge from the recombination of ideas across domains and engagement with unfamiliar concepts \citep{muthukrishna2016innovation,duede2024being}.

As LLMs are increasingly deployed in reasoning-intensive settings \citep{valmeekam2023planning,chen2024large}, preserving reasoning diversity becomes crucial. If these systems reflect only a narrow slice of human thought, they risk reinforcing dominant cognitive styles while marginalizing others. While LLMs offer high performance and access to expertise, collective convergence on even optimal algorithms can reduce decision-making quality in complex systems \citep{kleinberg2021algorithmic}. This concern is not abstract: a global ``weirdization'' may already be underway, as once-local ways of conceptualizing time, space, and causality give way to homogenized, Western-aligned models \citep{cooperrider2019happens}. Trained on massive, biased corpora and built upon similar datasets and architectures, LLMs may further accelerate this shift, risking the homogenization of reasoning processes and decision outcomes across contexts \citep{bender2021dangers,bommasani2022picking}.

{\color{blue}
Various studies have compared the reasoning and cognitive abilities of LLMs to those of humans across tasks traditionally used to assess human cognition. While these models often align with human reasoning in their outcomes, they have also revealed important discrepancies \citep{binz2023using,aher2023using,hagendorff2022thinking}. LLMs tend to produce reasoning patterns that cluster around central tendencies, lacking the natural variability that characterizes human thought. For example, in tasks inspired by the wisdom of the crowds phenomenon, where the diversity of individual judgments allows groups to collectively approximate correct answers, LLMs instead converge on uniform, ``idealized'' responses, missing the variance that makes human reasoning adaptive and collectively effective \citep{aher2023using}.
}

This mismatch may stem from the very objectives used to train and evaluate LLMs, which emphasize measurable performance gains or compliance with verifiable behavioral metrics such as accuracy, informativeness, helpfulness, harmlessness, or formatting consistency \citep{bai2022training,guo2025deepseek}, prioritizes improving reasoning performance over assessing variation in reasoning approaches. Even cognitively inspired methods, such as analogical reasoning in LLMs \citep{yasunaga2023large}, tree-of-thought prompting \citep{yao2023tree}, and memory-based inference \citep{wiratunga2024cbr}, are typically optimized for correctness and general utility. The widespread success of chain-of-thought prompting \citep{wei2022chain} has further reinforced this narrow focus,  driving homogenization across multimodal reasoning \citep{shao2024visual,han2025videoespresso}. This emphasis on performance, and the overreliance on techniques that maximize it, raises concerns about the adaptability, generalizability, and cognitive diversity of a uniform reasoning paradigm \citep{ziabari2025reasoning}. 
{\color{blue}
This overreliance and homogenization of reasoning already show documented downsides \citep{sui2025stop}. For instance, in a vehicle-classification task where most items followed a simple rule but a few violated it, chain-of-thought prompting made GPT-4o four times slower to learn correct labels, as step-by-step reasoning overgeneralized from regular patterns and overlooked exceptions and contextual cues \citep{liu2024mind}.
}

{\color{blue}
This imperative to maintain diversity extends beyond how LLMs reflect human reasoning to how they influence it. For instance, in controlled experiments designed to assess creative ideation, participants who received assistance from LLMs (i.e., ChatGPT) generated a greater number of more detailed and elaborated ideas, particularly benefiting those who were less experienced or less creative writers. However, their outputs were also judged to be more semantically similar across participants \citep{anderson2024homogenization,doshi2024generative}.
}
This convergence becomes even more concerning when considering how users interact with these systems: rather than actively steering generation, users often defer to model-suggested continuations, selecting options that seem ``good enough'' instead of crafting their own, which gradually shifts agency from the user to the model \citep{dang2023choice}. This tendency is particularly alarming for younger users, as LLMs' negative effects on creativity are most pronounced when used early in the ideation process \citep{qin2025timing}. Moreover, the consequences extend beyond creative output to the underlying cognitive processes: neurocognitive evidence shows that LLM-assisted writing elicits the weakest overall neural coupling, with reduced engagement of alpha and beta networks, lower memory recall, and diminished ownership of written work compared not only to independent writing but also to writing supported by search engines \citep{kosmyna2025your}.

\section{Concluding remarks}

Technological advancements have long reshaped human life, but LLMs stand out for the unprecedented scale and subtlety of their influence. 
Trained on vast and often biased corpora, biases further amplified through iterative retraining, LLMs are now deeply embedded in human language, intertwined with identity and cognition, and as billions interact with them on a daily basis, they risk homogenizing human expression and thought. Empirical evidence throughout this paper underscores this effect: reduced stylistic and lexical diversity in generated texts, subtle recalibration of user attitudes and framing in AI-mediated communication, and diminished diversity in ideation.

{\color{blue}
The concern is not just that LLMs shape how people write or speak, but that they subtly redefine what counts as credible speech, correct perspective, or even good reasoning by making certain framings more salient. As sociologist George Ritzer's ``McDonaldization'' theory suggests \citep{ritzer2021mcdonaldization}, processes favoring efficiency, predictability, and control can suppress contextual richness. LLMs mirror this logic in cognition, offering fluency and consistency while displacing situated, idiosyncratic forms of thought.
}

{\color{blue}
The centralized control over the very algorithms and datasets driving this homogenization is also acutely political, amplified by the dominance of a few platforms owned by multibillion-dollar corporations with significant political influence \citep{wiggers2025chatbots}. In a time of rising global populism, this concentrated power enables top-down homogenization, where perspectives can be subtly engineered for concentrated benefit and power. The documented censorship, such as the Chinese Qwen model's refusal to answer politically sensitive questions \citep{lin2024chinese_llm_censorship}, demonstrates this systemic risk. This algorithmic and market-driven erasure of diverse thought poses a modern danger akin to the linguistic control of Newspeak in Orwell's Nineteen Eighty-Four \citep{orwell1949nineteen}.
}

A growing number of strategies aim to counteract this homogenization, including personalized modeling approaches \citep{sorensen2024roadmap,aroyo2023dices}, embodied reasoning frameworks that promote situated and context-sensitive inference \citep{salaliflourishing}, and diversified prompting or multi-agent debate systems designed to broaden reasoning \citep{hu2024debate,du2023improving}. Yet evidence from sociolinguistics and psychology continues to document persistent homogenization, suggesting that these solutions must be applied more comprehensively and systematically before their effectiveness can be meaningfully evaluated. More broadly, both prompting- and training-based diversification methods remain constrained by underlying pretraining representations, potentially limiting their ability to induce deeper variation \citep{lin2023unlocking,wolf2023fundamental,ghosh2024closer}, and may also introduce trade-offs, as variation too far from pretraining distributions can increase the likelihood of hallucination \citep{gekhman2024does}.

In summary, while LLMs offer significant advancements and conveniences, their broad adoption without critical evaluation risks fundamentally altering the diverse cognitive landscapes that enrich human interaction and drive innovation. 
{\color{blue}
This homogenization constitutes a profound challenge to the ideal of collective wisdom, the fundamental principle, articulated by Mill, that true knowledge is acquired only by ``hearing what can be said about it by persons of every variety of opinion.''
We do not advocate for diversity in language, perspective, or reasoning without recognizing the coordination costs it can entail. Rather, we call for a deeper understanding of how LLMs affect the diversity of language and thought, as well as the signals embedded in this diversity, and for these insights to inform their design. Moving forward, preserving and enhancing meaningful human diversity should be a central criterion in the development and evaluation of LLMs.
}
Only through deliberate attention to this pluralism can we harness the full potential of language technologies without sacrificing the very diversity that defines human society.
Nevertheless, there is still much to learn about the LLM‑driven homogenization of language and thought, and how best to address it. Some key directions for deepening this understanding are outlined in the section below.
\clearpage

\section{Outstanding Questions}

\begin{itemize}[leftmargin=1.5em]
    \setstretch{1.2}

\item Will current alignment methods, such as supervised fine-tuning and RLHF, ever be sufficient to reproduce the full diversity of human cognition, or are more foundational changes in model architecture, objectives, and training data required?
These approaches have increased model steerability and surface-level variation, but it remains unclear whether they can capture deeper context-sensitive and culturally grounded forms of diversity found in human thought.

\item Even if LLMs produce more diverse outputs, how can we ensure that this diversity is meaningful and grounded in actual human experience rather than being superficial or artificially constructed?
Future research should identify and promote metrics and frameworks that distinguish synthetic variation from diversity reflecting authentic sociocultural, emotional, and cognitive nuance.

\item What are the long-term cognitive effects of sustained reliance on LLMs for ideation, writing, and reasoning?
While short-term effects such as reduced stylistic variation and creative ownership have been observed, we lack longitudinal studies that track changes in abstraction, memory retention, and reasoning strategies over time and examine whether changes are irreversible.

\item Can users be equipped with strategies to counteract the homogenizing effects of LLMs on expression and thought?
Research is needed to develop and evaluate behavioral or interface-level interventions, such as delaying LLM use during ideation, or exposing users to model-induced changes, that help preserve agency and individuality in interaction.

\item What taxonomies or repertoires of intervention can future research establish to help mitigate LLM-driven homogenization at scale?
A systematic understanding of the possible safeguards, behavioral, architectural, or institutional, is needed to guide users, developers, and platforms toward practices that promote cognitive and linguistic pluralism.

\end{itemize}
\clearpage

\section*{Glossary}


\textbf{Chain-of-Thought (CoT) prompting:} A prompting strategy that encourages models to show step-by-step reasoning before giving an answer, improving structure and accuracy but sometimes reducing intuitive or creative responses.  


\textbf{Epistemic:} Relating to knowledge, understanding, or how beliefs are formed and justified. Used to describe cognitive or informational aspects of learning and reasoning.  

\textbf{Epistemic Collapse:} A situation where diversity in knowledge, reasoning, or interpretation is lost, leading to uniform ways of thinking and a narrower collective understanding.  

\textbf{Essentialized Representations of Identity:} Simplified portrayals of social or cultural groups that treat identity as fixed and homogeneous, often ignoring variation and individual nuance.  



\textbf{Natural Language Processing (NLP):} A field of AI focused on enabling computers to understand, interpret, and produce human language across tasks like translation, summarization, and question answering.  


\textbf{Prompt:} A text input or instruction given to a language model that guides how it generates a response, effectively shaping the model's behavior and output.

\textbf{Reinforcement Learning from Human Feedback (RLHF):} A training method where human evaluators rate model outputs, guiding the model to produce more preferred or context-appropriate responses.  

\textbf{Supervised Fine-Tuning:} A supervised training process that adapts a pre-trained language model to a specific task or domain by continuing its training on a smaller, labeled dataset, teaching the model new knowledge or behaviors.

\textbf{Temperature Scaling:} A parameter that controls how deterministic or random a model's outputs are: lower temperatures yield precise, predictable text, while higher ones increase variety.

\section*{Acknowledgments}
This research was supported by the Air Force Office of Scientific Research A9550-23-1-046. The funders had no role in study design, data collection and analysis, decision to publish, or preparation of the manuscript. The views and conclusions contained herein are those of the authors and should not be interpreted as necessarily representing the official policies, either expressed or implied, of AFOSR, or the U.S. Government. The U.S. Government is authorized to reproduce and distribute reprints for governmental purposes notwithstanding any copyright annotation therein.

\bibliographystyle{unsrt}
\bibliography{llm_diversity_refs}

\end{document}